\documentclass[sigconf,acmthm]{acmart}
\usepackage{tikz}
\usepackage{tabularx}
\usepackage{multirow}
\usepackage{hhline}
\usepackage{amsmath}
\usepackage{todonotes}
\usepackage{tikz}
\usepackage{graphics}
\DeclareRobustCommand{\stirling}{\genfrac\{\}{0pt}{}}

\usetikzlibrary{patterns,arrows,decorations.pathreplacing}

\def\BibTeX{{\rm B\kern-.05em{\sc i\kern-.025em b}\kern-.08emT\kern-.1667em\lower.7ex\hbox{E}\kern-.125emX}}
    
\copyrightyear{2019}
\acmYear{2019}
\acmConference[RecSys '19]{Thirteenth ACM Conference on Recommender Systems}{September 16--20, 2019}{Copenhagen, Denmark}
\acmBooktitle{Thirteenth ACM Conference on Recommender Systems (RecSys '19), September 16--20, 2019, Copenhagen, Denmark}
\acmPrice{15.00}
\acmDOI{10.1145/3298689.3347006}
\acmISBN{978-1-4503-6243-6/19/09}
\setcopyright{rightsretained}

\begin{document}

\title{A Deep Learning System for Predicting Size and Fit in Fashion E-Commerce}
\author{Abdul-Saboor Sheikh}
\email{saboor.sheikh@zalando.de}
\affiliation{Zalando Research}
\author{Romain Guigoures}
\authornote{Work done while at Zalando SE}
\email{romain.guigoures@shopify.com}
\affiliation{Shopify}
\author{Evgenii Koriagin}
\email{evgenii.koriagin@zalando.de}
\affiliation{Zalando SE}
\author{Yuen King Ho}
\authornotemark[1]
\affiliation{OLX Berlin}
\author{Reza Shirvany}
\email{reza.shirvany@zalando.de}
\affiliation{Zalando SE}
\author{Roland Vollgraf}
\email{roland.vollgraf@zalando.de}
\affiliation{Zalando Research}
\author{Urs Bergmann}
\email{urs.bergmann@zalando.de}
\affiliation{Zalando Research}


\renewcommand{\shortauthors}{Sheikh, et al.}

\begin{abstract}
Personalized size and fit recommendations bear crucial significance for any fashion e-commerce platform. Predicting the correct fit drives customer satisfaction and benefits the business by reducing costs incurred due to size-related returns. Traditional collaborative filtering algorithms seek to model customer preferences based on their previous orders. A typical challenge for such methods stems from extreme sparsity of customer-article orders.
To alleviate this problem, we propose a deep learning based content-collaborative methodology for personalized size and fit recommendation. 
Our proposed method can ingest arbitrary customer and article data and can model multiple individuals or intents behind a single account. The method optimizes a global set of parameters to learn population-level abstractions of size and fit relevant information from observed customer-article interactions. It further employs customer and article specific embedding variables to learn their properties. Together with learned entity embeddings, the method maps additional customer and article attributes into a latent space to derive personalized recommendations. 
Application of our method to two publicly available datasets demonstrate an improvement over the state-of-the-art published results. On two proprietary datasets, one containing fit feedback from fashion experts and the other involving customer purchases, we further outperform comparable methodologies, including a recent Bayesian approach for size recommendation.
\end{abstract}



\maketitle

\section{Introduction}
\label{intro}
Fashion is a way to express identity, moods, and opinions. Recent studies show size and fit are amongs the most influential factors, driving e-commerce customer satisfaction \cite{Pisut2017}. A crucial difference when engaging in online compared to traditional brick and mortar retail is the lack of immediate sensory feedback about fit and feel of a product. For many, this is a major deterrent against fashion e-commerce. 

To make matters worse, the notion of size is inherently ambiguous: for instance, size systems may be coarsely defined (e.g `Small' , `Medium', `Large' ) or they may vary between regions (e.g., EU vs. US shoe sizes). There is furthermore vanity sizing, where brands modify standardized size specifications to target a particular clientele. As a result, there exists myriad of overlapping size systems in the fashion industry, with no agreed standard for conversion between them. Even within brands there is not necessarily one consistent conversion logic employed to convert sizes from one country or region to another.

One way to assist customers in finding the correct size is to provide size conversion charts which convert body measurements to article sizes. However, this requires customers to know their body measurements. Interestingly, even if the customer gets accurate measurements with the aid of tailor-like tutorials and expert explanations, the size charts themselves almost always suffer from high variance, even within a single brand. This is especially true for fast fashion brands that represent the largest part of sales volume. In a fast moving fashion environment, designers strive to beat competition by continuously serving consumers with the latest trends at competitive prices. To meet time, cost and design constraints, same articles with varying attributes (e.g., color, material, etc.) are often sourced from different production channels, causing inconsistencies in size and fit characteristics. 

There are numerous other factors that make it essential for fashion e-commerce platforms to develop data-driven systems for providing informed size and fit advice to their customers~\citep[e.g.,][]{abdulla2017, guigoures2018hierarchical, sembium2017, Sembium2018, misra2018decomposing}.

In this work, we propose a deep learning based content-collabora\-tive methodology for personalized size and fit prediction. Standard approaches to collaborative filtering solely rely on interaction data to model customer behavior~\cite{KorenBell2015}, but for a vast majority of customers, such data is sparse. This results in an extremely sparse customer-article interaction matrix, which makes it difficult to model preferences of every individual customer on a personalized level. Additional information in the form of customer and article attributes can however help to deal with the sparsity and cold-start recommendations~\citep[see e.g.,][]{ShiEtAl2014,sembium2017}. In the same spirit, our proposed method uses both interaction data as well as arbitrary customer and article features for personalized size/fit prediction. Our method employs a split-input neural network architecture with global and entity-specific parameters. The global set of parameters allows the model to capture information relevant for predicting size and fit across customers, whereas the entity-level embedding variables equip the model with the capacity to discover implicit properties of individual customers and articles for personalized recommendations. The method is a priori independent of underlying semantics behind its targets and can model multiple individuals or intents behind an account.

\section{Related Work}
\label{sec:2}

The topic of understanding article size issues as well as predicting size and fit on a personalized level has gained momentum in the research community. In the following we outline some recent developments on the subject and draw parallels between our work and closely-related methodologies in collaborative filtering:

The authors of~\cite{PengSayegh2014} put forth the idea of mapping customer images to existing 3D body scans, which are aligned with articles to generate fit ratings.

The method introduced in~\cite{abdulla2017} proposes to use a skip-gram based word2vec model~\cite{word2vec} on the purchase history data to learn latent representations of articles. The approach then forms a customer representation by aggregating over the learned representations of said customers' purchased articles. A gradient boosted classifier is then trained on customer and article latent representations to predict the fit. 

In ~\cite{guigoures2018hierarchical}, the authors propose a hierarchical Bayesian approach for personalized size recommendation. Conditioned on customer and article pairs, the method models the joint conditional probability of sizes ordered by customers together with their outcomes (i.e. kept vs. size related return) as observed in training data. For making personalized size recommendations, the method uses the conditional probability of size given a customer and an article with the outcome set to keep. The method uses approximate probabilistic inference for parameter optimization and testing.    

The authors of \cite{sembium2017} propose to deduce `true' sizes of customers and articles from purchase and return data using a latent factor model. The deduced size features are fed into a standard classification regime to perform ordinal fit prediction (i.e. `Small', `Fit', `Large'). The method in addition performs hierarchical clustering on individual customer data to handle multiple customers behind an account. A follow-up work proposes a Bayesian version of the ordinal regression model \cite{Sembium2018}. The method relies on approximate probabilistic inference (mean-field variational approximation with Polya-Gamma augmentation) for posterior distribution estimation over customer and article sizes.  

An approach conceptually similar to our work is proposed in \cite{misra2018decomposing}, which models the size recommendation problem as a fit prediction problem. In a two-step procedure, the method first learns to embed customers and articles in a latent space with the same dimensionality. Once the embeddings are obtained using an ordinal regression procedure, they are used in the next step to learn representations for each class by applying prototyping and metric learning techniques. The authors of \cite{misra2018decomposing} also provide the public datasets that we use to benchmark our approach.

\begin{figure}
\begin{center}
\includegraphics[width=0.88\columnwidth]{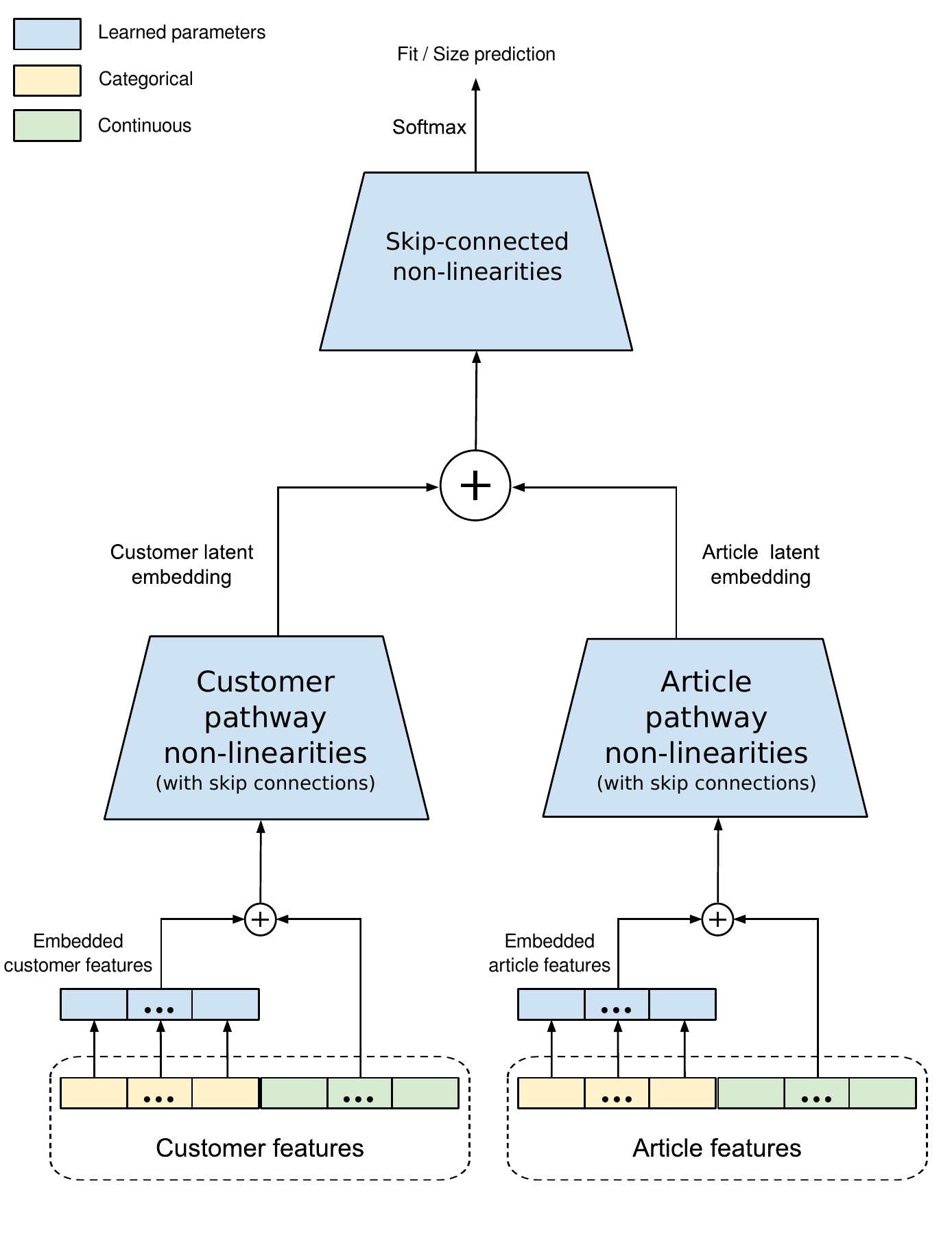}
\caption{Schematic of \textsc{SFnet} architecture for size and fit prediction. The $\oplus$ symbol indicates concatenation, while each trapezoid represent a cascade of fully-connected feedforward layers with skip connections.}
\label{fig:nn_architecture}
\end{center}
\end{figure}

Most of the works mentioned above do not take an end-to-end approach to the task at hand, while some are limited w.r.t. scalablility (e.g., due to their probabilistic nature) or capacity (e.g., due to predefined interactions, linearity assumptions, ability to handle cold-starts or model multiple users/intents behind one identity). Our work in contrast presents a scalable, end-to-end deep learning approach to size and fit recommendation. The two pathway neural network architecture employed in this work (Figure~\ref{fig:nn_architecture}) flexibly consumes both categorical and continuous customer and article features and it learns (potentially non-linear) customer-article interactions from data. 

Our model architecture is rather generic in the context of collaborative filtering. It is for instance closely related to the Deep Structured Semantic Model (DSSM)~\cite{DSSM2013} and Neural Collaborative Filtering (NCF)~\cite{ncf2017}. 
Developed for web search, DSSM uses independent neural network layers to embed customers and articles into a latent space. It then uses a predefined interaction between the latent embeddings to predict its target. 
NCF employs a Neural Tensor Networks~\cite{socher2013} inspired architecture to learn input embeddings or features for (one-hot encoded) customers and articles. The architecture comprises a shallow (GMF) as well as a deep (MLP) feedforward pathway
to respectively model both linear and non-linear interactions between customer and article pairs. A notable difference between our architecture and DSSM or NCF is that our architecture uses skip connections~\cite{resnet2016} between layers.

Our proposed approach can be seen as a generalization of logistic matrix factorization~\cite{johnson2014logistic}, which is a linear model of customer-item interactions. Aside from interaction data, the method does not take any additional customer or item information into account for making personalized recommendations.

\section{Problem Formulation}
\label{sec:3}
We build our recommendation system via likelihood maximization. 
To that end, we ought to formulate and optimize the parameters of an instance of a probabilistic model that maximizes the probability of 
outcomes of observed customer-article interactions in the training data. 
Our training data is a set of $N$ tuples $\mathcal{D} = \{c,a,o\}_{n=1,\ldots,N}$, where $c$ denotes a customer, $a$ an article and $o$ is a categorical variable such as fit feedback or size of the article. Given the data we can define a conditional probability distribution $p(o\mid c,a)$, such that it allows  
us to define a statistical model for associating customer-article interactions with respective outcomes. Given $p(o\mid c,a)$ and a set of $N$ customer-article interactions, we can define the following likelihood function:

\begin{equation}
    \mathcal{L}(\Theta,\mathcal{D}) = \prod_{n=1}^N p(o^{(n)} \mid c^{(n)},a^{(n)};\Theta),  
    \label{EqLkh}
\end{equation}

where $\Theta$ represents the set of parameters of the conditional distribution. We seek values for $\Theta$ so that (\ref{EqLkh}), or equivalently its logarithm, is maximized. 
Once optimized, we can evaluate the conditional distribution with customer-article pairs to estimate the odds of modeled outcomes, i.e. size or fit. For brevity, we will omit $\Theta$ in our later references to the conditional distribution in (\ref{EqLkh}).

\subsection{Modeling Assumptions}
\label{subsec:1}
In (\ref{EqLkh}) we make a simplifying assumption that each of the $N$ data points in the training dataset is independently and identically distributed given a customer and article pair. This allows us to model the outcome $o$ as a categorical variable. One can however consider modeling $o$ as a multivariate categorical vector $\vec{o}$ e.g., to capture interactions among multiple sizes in selection-orders -- orders where a customer orders more than one sizes.
Such a modelling scheme would allow to capture co-dependencies among the elements of $\vec{o}$, but at the cost of increased model complexity.  

Furthermore both this work and other models compared here do not take the temporal nature of the data into account. A more elaborate model could further condition every order on all previous orders. 

As we shall see, the simplifying assumptions discussed above yield a computationally amenable objective (\ref{EqLkh}) that can be optimized at scale in an end-to-end fashion for predicting customer size or fit on a personalized level for a given query article.

\subsection{Modeling Personalized Size/Fit Preferences}
\label{subsec:2}
In general, the conditional distribution in (\ref{EqLkh}) takes the form of a categorical distribution over one of $k$ possible outcomes of the output variable $o$. 
For instance, in case of a binary outcome (e.g., $o \in \{$`Fit', `No fit'$\}$), $p(o \mid .)$ can be modeled as a Bernoulli distribution. In the simplest form, we can marginalize over all the articles in a customer's history to have $p(o \mid c)$ only conditioned on the customer. Such a customer-only-level personalization approach (with some population-level smoothing) aggregates over articles, and hence to a certain degree alleviates the data sparsity problem. Marginalization of articles may also be a reasonable assumption so long as customers size 
and fit preferences are not influenced by article attributes. However, article attributes, including brand, style, material etc. can indeed influence a customer's size preferences, which makes it desirable to model dependencies of such kind even when individual customer order histories may only sparsely reflect such fine-grained information. We therefore define a global model of $p(o \mid c,a)$ such that its parameters are (partially) shared across all customers and articles:

\begin{align}
\label{EqConDist}
  p(o \mid c,a) &=  \text{Categorical}(\vec{\omega}), \\
  \mbox{where}\ \ \ \ \vec{\omega} = N&N(\psi_c,\psi_a; \mathcal{W}),\  \ \ \ s.t. \ \ \ \ \sum_k \vec{\omega}_k = 1 . \nonumber 
\end{align}
Here we define the parameters of $p(o\mid c,a)$ to be the output of a neural network (i.e. $\vec{\omega}$ is the $\mathrm{softmax}$ output of a feedforward neural network). The elements of the vector $\vec{\omega}$ signify the odds of $k$ possible 
outcomes such as sizes of an article or one of the $k$ possible fit feedback values.   
Our neural network is parameterized by a set of matrices $\mathcal{W}$ and consumes feature sets $\psi_c$ and $\psi_a$ corresponding to both customer and article.
The features can be comprised of both explicit attributes as well as variables that can be uniquely identified with individual customers and articles and allow us to encode implicit information such as customer style preferences or intrinsic article sizes. As we will see in Section \ref{subsec:3}, such 
encodings in neural network based models can be learned in an end-to-end fashion by means of input feature embeddings. 

By plugging (\ref{EqConDist}) into (\ref{EqLkh}), we can globally optimize for $\Theta$  by minimizing a loss function such as categorical cross-entropy via (stochastic) gradient descent (SGD). Note that  $\Theta$ includes neural network weight matrices $\mathcal{W}$ as well as the embedded input features of customers and articles.

\subsection{Size and Fit Network (\textsc{SFnet}) Architecture}
\label{subsec:3}
For the neural network in  (\ref{EqConDist}), we choose an architecture that is loosely inspired by Siamese networks~\cite{BromleyEtAl1993}; however, there is a crucial difference that input pathways of the model are not weight sharing replica of each other~\cite{Elkahky2015}. As illustrated in Figure \ref{fig:nn_architecture}, the size and fit network (\textsc{SFnet}) architecture ingests customer and article information through non-identical feedforward input pathways. As shown in the figure, the input layers of both customer and article pathways embed categorical features (e.g., customer id, article id, brand, etc.) such that their unique values get mapped to trainable vector variables. Note that by embedding unique customer or article identifiers, we indeed equip the model with the capacity to learn personalized latent features of individual customers and articles in an end-to-end fashion. Both customer and article input pathways concatenate their set of embedded and non-embedded (i.e. continuous) features to pass them through a cascade 
of non-linear layers with skip connections~\cite{resnet2016} to obtain latent embeddings of customers and articles. This allows the model to capture latent information about both entities that is only contained in (higher-order) implicit patterns in data.  Through such an embedding scheme, we can theoretically learn to disentangle information 
and identify multiple personas with diverging size or fit preferences behind a single account or discover properties that are intrinsic to certain articles or brands.

After obtaining the so called latent embeddings of both customer and article, we simply concatenate the embeddings to send the combined information through another set of non-linearities (with skip connections) to yield the parameter vector $\vec{\omega}$ which paramterizes the conditional disrtibution (\ref{EqConDist}). 

In the neural network architecture described above, the continuous features as well as the learned input embeddings of categorical features jointly allow the model to represent customers and articles on a personalized level. On the other hand, through the weight matrices $\mathcal{W}$ which paramterize the network layers, the model learns to represent higher-order patterns in the data that are globally relevant for predicting size and fit. Such a model can be efficiently trained at scale, given (individually) sparse customer-article interaction histories. 

\section{Empirical Evaluation}
\label{sec:4}
We demonstrate the generality of our method by applying it to different datasets and tackle a variety of size and fit related classification tasks. Two of the datasets we use are publicly available benchmarks for size recommendation \cite{misra2018decomposing}, while another two are our internal datasets. One of the internal datasets contains feedback from fashion experts on length and width deviation of a large number of shoes with respect to their given sizes. The other internal dataset is comprised of a large number of customer orders and purchases, on which in a backtesting setup we learn to predict sizes of ordered and kept articles for individual customer accounts. We compare our approach with a number of methodologies and report micro-averaged area under the ROC curve (AUC), accuracy and average log-likelihood as performance metrics.

\subsection{Experimental Setup}
\label{sec:2.1}
We use the Keras functional API with Tensorflow backend in Python for our implementation. For parameter optimization we use the Adam optimizer~\cite{kingma2014} to perform SGD. We use performance on validation data (taken to be a $10\%$ split of the data at hand) for hyperparameter tuning and to avoid overfitting. 
Table \ref{tab:hyperparams} describes the hyperparameter settings we used in our experiments. \footnote{The settings listed in Table \ref{tab:hyperparams} were not found exhaustively and in our experience the performance is fairly robust to minor deviations in the listed settings. Apart from $L2$ regularization as listed in Table \ref{tab:hyperparams}, we did not observe significant performance gains from applying other regularization measures such as dropout~\cite{srivastava2014} or batch normalization~\cite{ioffe2015}.} 

Due to the input embedding of categorical features, the parametric capacity and with it the memory requirement of our method increase linearly with respect to both the cardinality of embedded customer and article features, as well as customer and article numbers. Otherwise the number of parameters as defined by customer and article input pathways and top layers in Table \ref{tab:hyperparams} remains constant throughout.

\begin{table}[!ht]
\caption{Hyperparameter settings used in our experiments.}
\label{tab:hyperparams}     
\resizebox{\columnwidth}{!}{
\begin{tabular}{ll}
\hline\noalign{\smallskip}
\textsc{SFnet} Hyperparameters &  \\
\noalign{\smallskip}\hline\noalign{\smallskip}
Customer/Article Pathway &   {$\#$ (emb. + cont.) feats. $\times$ 25 $\times$ 15 $\times$ 10} \\
Top Layers &     {50 $\times$ 100 $\times$ 200 $\times$ 500 $\times$ $\mathrm{softmax}$ output }  \\
L2 Reg. $\mathcal{W}$ &  {--}  \\
L2 Reg. Cust. Emb. & {0.1}  \\
L2 Reg. Article Emb.& {0.01} \\
Embedding Dimensions& {10}  \\
Hidden Unit Activation &{tanh}  \\
Loss & {cross-entropy} \\
SGD Batch Size& 2048  \\
Epochs & 15--50 \\
\noalign{\smallskip}\hline
\end{tabular}
}
\end{table}

\subsection{Experiments on Public Datasets}
\label{sec:2.2}
The two publicly available datasets we use were introduced by \cite{misra2018decomposing}. One of the datasets `ModCloth' comes from an online vintage clothing retailer. The data contains three categories 
of clothing: dresses, bottoms and tops. The other dataset `RentTheRunWay' comes from an online clothing rental platform for women. The dataset is comprised of several clothing categories (including shoes). Both datasets contain customer-article interactions with categorical feedback on fit: `Small', `Fit' or `Large'. 
Table \ref{tab:datastats} contains general statistics of the datasets as provided by \cite{misra2018decomposing}. The datasets are sparse in customer-article interaction. Following the protocol used by \cite{misra2018decomposing}, we randomly split the data into $80\%$ training, $10\%$ validation and $10\%$ testing; however, since we do not know the exact split used in \cite{misra2018decomposing}, we report the average results with standard deviation computed from $10$ independent trials.

\begin{table}
\centering
\caption{General statistics of public datasets.}
\label{tab:datastats}      
\begin{tabular}{lll}
\hline\noalign{\smallskip}
Statistic/Dataset & ModCloth & RentTheRunWay  \\
\noalign{\smallskip}\hline\noalign{\smallskip}
\# Transactions & 82,790 & 192,544 \\
\# Customers & 47,958 & 105,571 \\
\# Articles & 5,012 & 30,815 \\
\% Small & 15.7 & 13.4 \\
\% Large & 15.8 & 12.8 \\
Single Transaction Customers& 31,858 & 71,824 \\
Single Transaction Articles & 2,034 & 8,023 \\
\noalign{\smallskip}\hline
\end{tabular}
\end{table}

Table \ref{tab:benchmarkfeats} lists customer and article features available in both datasets that we use to train our neural network. We indicate further categorical features we embed via the input embedding technique described in Section \ref{subsec:2}. To handle cold-start cases during test (and validation), we define a `default' input embedding for each embedded feature. The default embeddings were then trained by randomly and independently assigning each of them, $10\%$ of the data points every SGD epoch.

\begin{table*}
\centering
\caption{Benchmark customer and article features. 
Features marked with $^*$ were categorical and were embedded using input embedding. Moreover, 
features markded with $^+$ were split into alphabetical (for embedding) and numerical parts.}
\label{tab:benchmarkfeats}       
\begin{tabular}{lll}
\hline\noalign{\smallskip}
Features/Dataset & ModCloth & RentTheRunWay  \\
\noalign{\smallskip}\hline\noalign{\smallskip}
Article & category$^*$, quality, item id$^*$, size & category$^*$, rating, rented for$^*$, item id$^*$, size\\ \noalign{\smallskip}
 Customer & shoe width$^*$, shoe size, waist, bust, cup size, bra size, & age, body type$^*$, bust size$^+$, height, weight, user id$^*$ \\
 & hips, height, user id$^*$ & \\
\noalign{\smallskip}\hline
\end{tabular}
\end{table*}

\vspace{3mm}
\noindent{\textbf{MLP Baseline:}}
As a deep learning baseline, we train another neural network to parameterize (\ref{EqConDist}). The architecture of the model is a feedforward neural network that we obtain by simply concatenating the customer and article input pathways of \textsc{SFnet}. It therefore corresponds to the MLP pathway of NCF~\cite{ncf2017}, however with additional customer and article input features and skip connections between layers. For both benchmarks, the network takes as input a concatenated set of customer and article features listed in  Table \ref{tab:benchmarkfeats}. All categorical features marked in the table are embedded via input embeddings. We follow hyperparamter settings from Table \ref{tab:hyperparams} to endow the model with a capacity comparable to \textsc{SFnet}. Following the same protocol as for \textsc{SFnet}, we perform $10$ independent runs of the model to report mean and standard deviation of the performance metrics.

\begin{figure*}
\begin{center}
\includegraphics[width=0.69\textwidth]{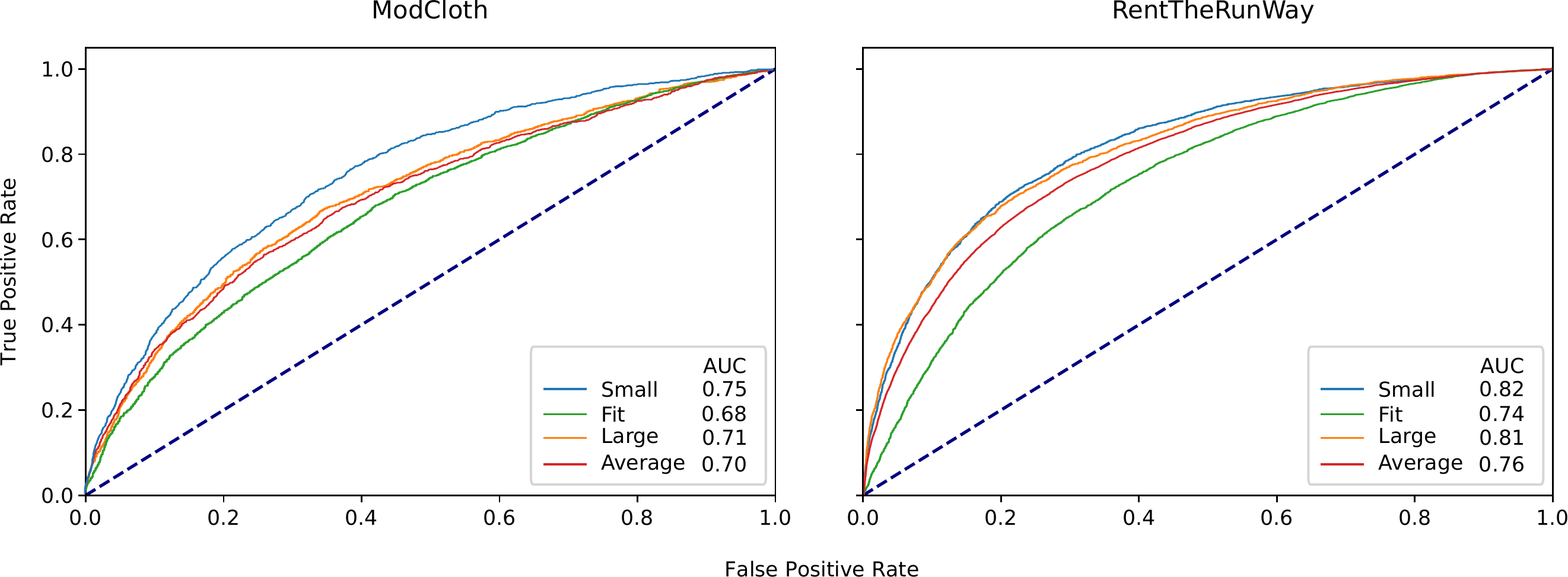}
\caption{The ROC curves for one of the best runs of \textsc{SFnet} on benchmark datasets.}
\label{fig:benchmarks}
\end{center}
\end{figure*}

\vspace{3mm}
\noindent{\textbf{Results:}}
We compare the performance of \textsc{SFnet} on benchmark datasets in Table \ref{tab:benchmarkperf}. The first four rows in the table are results from \cite{misra2018decomposing}, where the authors compare latent variable (LV) vs. latent factor (LF) based embeddings of customers and articles with logistic regression (LR) or metric learning (ML) on top for classification. The approach is conceptually analogous to ours, but we learn both customer and article embeddings as well as their interaction end-to-end with a neural network. To our knowledge, the results of \cite{misra2018decomposing} represent the previous state-of-the-art on both benchmarks; \textsc{SFnet} however clearly outperforms~\cite{misra2018decomposing} as well as the MLP baseline, is analogous to the MLP pathway in NCF. 
As illustrated in Figure \ref{fig:benchmarks}, in one of our runs we could achieve more than $5\%$ improvement on the average AUC over the previously best performing LF-ML. While~\cite{misra2018decomposing} do not publish results on accuracy and average log-likelihood, compared to the MLP baseline, \textsc{SFnet} achieves better results on both datasets.

\begin{table*}
\centering
\caption{Comparison on publicly available Benchmark datasets.}
\label{tab:benchmarkperf}      
\begin{tabular}{lllllll}
\hline\noalign{\smallskip}
 &  \multicolumn{2}{c}{Micro-avg. AUC} &   \multicolumn{2}{c}{Accuracy} & \multicolumn{2}{c}{Average log-likelihood}  \\
 \noalign{\smallskip}\hline\noalign{\smallskip}
Method/Dataset & ModCloth & RentTheRunWay & ModCloth & RentTheRunWay & ModCloth & RentTheRunWay   \\
\noalign{\smallskip}\hline\noalign{\smallskip}
LV-LR & 0.617 & 0.676 &  -- & -- & -- & -- \\
LF-LR & 0.626 & 0.672 &  -- & -- & -- & -- \\
LV-ML & 0.621 & 0.681 &  -- & -- & -- & -- \\
LF-ML & 0.657 & 0.719 &  -- & -- & -- & -- \\
MLP Baseline &  0.624 $\pm$ 0.007 & 0.692 $\pm$ 0.010 & 0.681 $\pm$ 0.004 & 0.733 $\pm$ 0.006 & -0.819 $\pm$ 0.004 & -0.708 $\pm$ 0.01 \\
\textsc{SFnet} & \bf 0.689 $\pm$ 0.005 & \bf 0.749 $\pm$ 0.004 & \bf 0.690 $\pm$ 0.004 & \bf 0.760 $\pm$ 0.004 & \bf -0.758 $\pm$ 0.006 & \bf -0.610 $\pm$ 0.008 \\
\noalign{\smallskip}\hline
\end{tabular}
\end{table*}

\subsubsection{\textbf{Customer and Article Embeddings and Data Sparsity:}}
As discussed in Section \ref{subsec:3}, the method we propose can learn implicit features of customers and articles through entity-specifc input embeddings; the model however requires enough interactions of an entity (i.e. a customer or an article) to learn its meaningful representation through input embedding. This is evident in Table \ref{tab:ablemb}, where we compare the performance of \textsc{SFnet} on ModCloth and RentTheRunWay benchmarks w.r.t. inclusion vs. exclusion of user and item identifiers from customer and article features. As indicated by the first two rows of the table, we observe including or excluding user ID from the list of customer features in Table \ref{tab:benchmarkfeats} does not have a significant effect on performance for both the datasets. This should not come as a surprise as the general statistics of data in Table \ref{tab:datastats} indicate that most customers in both datasets have only one transaction, hence we cannot expect the model to capture anything meaningful by embedding the customer identifier. Table \ref{tab:datastats} on the other hand indicates that the datasets are relatively sparse on the article side. Indeed removing item ID from article features in Table \ref{tab:benchmarkfeats} affects the performance of our model, which is reflected by the third and fourth rows of Table \ref{tab:ablemb}.

Given these results for the benchmarks, we surmise that \textsc{SFnet} makes use of both explicit and implicit features of articles, while for customers it mainly relies on explicit features to handle the task. In the next sections, our method will completely rely on input embeddings learned against unique identifiers to represent customers for personalized size and fit predictions.

\begin{table*}
\centering
\caption{Effect of including or excluding customer and article embeddings on the performance of \textsc{SFnet}.}
\label{tab:ablemb}      
\begin{tabular}{llllllll}
\hline\noalign{\smallskip}
\multicolumn{2}{c}{Entity embedding} & \multicolumn{2}{c}{Micro-avg. AUC} & \multicolumn{2}{c}{Accuracy} & \multicolumn{2}{c}{Average log-likelihood}  \\
 \noalign{\smallskip}\hline\noalign{\smallskip}
user id & item id & ModCloth & RentTheRunWay & ModCloth & RentTheRunWay & ModCloth & RentTheRunWay   \\
\noalign{\smallskip}\hline\noalign{\smallskip}
\checkmark & \checkmark & \bf 0.689 $\pm$ 0.005 & \bf 0.749 $\pm$ 0.004 & \bf 0.690 $\pm$ 0.004 & \bf 0.760 $\pm$ 0.004 & \bf -0.758 $\pm$ 0.006 & \bf -0.610 $\pm$ 0.008 \\
 $\times$ & \checkmark & \bf 0.693 $\pm$ 0.009 & \bf 0.751 $\pm$ 0.004 & \bf 0.691 $\pm$ 0.004 & \bf 0.760 $\pm$ 0.001 & \bf -0.757 $\pm$ 0.009 & \bf -0.607 $\pm$ 0.004 \\
\checkmark & $\times$ &  0.637 $\pm$ 0.004 & 0.667 $\pm$ 0.007 & 0.686 $\pm$ 0.004 & 0.733 $\pm$ 0.007 & -0.803 $\pm$ 0.006 & -0.716 $\pm$ 0.023 \\
$\times$ & $\times$ &  0.638 $\pm$ 0.007 & 0.674 $\pm$ 0.003 & 0.683 $\pm$ 0.005 & 0.739 $\pm$ 0.002 & -0.806 $\pm$ 0.009 & -0.698 $\pm$ 0.006 \\
\noalign{\smallskip}\hline
\end{tabular}
\end{table*}

\subsection{Experiments on Expert Feedback Data}
\label{sec:2.3}
In order to gain insights on size and fit characteristics of new articles before their online activation, we ask different fashion experts to physically try on articles and provide qualitative feedback on their fit. Each fitting session involves one fashion expert and the sessions are run independently so that the experts do not influence each other. We run three fitting sessions for each article. For every session we draw an expert from a pool of $55$ experts. 

The motivation for this experiment is that if using \textsc{SFnet} we can learn to reliably predict fit feedback of individual experts given the attributes of an article, we can select new articles for try-ons based on the predicted feedback: for instance when there is a degree of disagreement in the predicted feedback of different experts or if there is a consensus on deviation from true to size fit. 

The data for the experiment is comprised of around $30K$ distinct pairs of shoes. We collect feedback on both length and width of the shoes. The feedback is defined as an ordinal variable and it takes one of the $5$ values: `Too small', `Small', `True to size', `Big' or `Too big'. The dataset is highly imbalanced with $73.7 \%$ and $87.1 \%$ true to size responses for length and width.

We train individual instances of \textsc{SFnet} and the methods we compare with to independently predict the feedback on length and width. We treat each fashion expert as a customer who is represented by a unique identifier. For shoes we use attributes such as brand, fitted size, color, main material and $5$ other categorical attributes, which define non-overlapping subcategories of shoes. All features we consider are categorical and are embedded through input embedding. We perform $10$ independent runs to report the mean and standard deviation of the performance metrics.
For each run we randomly split data to consume $80\%$ for training, while $10\%$ each is kept for validation and testing. We benchmark our method against two other well-suited approaches for the problem: a Naive Bayes classifier and boosted trees. 

\vspace{3mm}
\noindent{\textbf{Naive Bayes:}}
When dealing with classification tasks with categorical input features, Naive Bayes is a straightforward choice. However in our case some of the features have very high cardinality (over $500$ distinct brands for example) and some of the feature values are sparsely or never observed in the training data. Hence we apply Laplace smoothing \cite{manning2013introduction} to avoid computational issues with the conditional probability estimation. 

\vspace{3mm}
\noindent{\textbf{Boosted trees:}} 
Another well-suited methodology to compare against is gradient boosted trees. High feature cardinality also poses a problem for tree based approaches as it requires the training algorithm to evaluate the best of all the possible partitions of $n$ feature values into $k$ classes, which is equal to the Stirling number of second kind $\stirling{n}{k}$ \cite{graham1989concrete}. We therefore encode fashion experts and shoe attributes using smoothed target encoding \cite{micci2001preprocessing} to reduce the complexity of the task. 

\vspace{3mm}
\noindent{\textbf{Results:}}
Table \ref{tab:expertfeedback} shows the results obtained on test data. All three approaches are comparable in terms of accuracy; however, the numbers hover around the a priori probability ($73.7 \%$ for length and $87.1 \%$ for width) of the dominant `true to size' class. We take the results as an indication of expert feedback being unbiased and therefore independent of the considered article attributes. In terms of other metrics, while~\textsc{SFnet} takes a clear lead w.r.t. the average AUC, the relatively low likelihood values of \textsc{SFnet} despite being more accurate in comparison to Naive Bayes suggests that the output distributions of \textsc{SFnet} may tend to be more peaky in nature. This leads to a relatively high loss in likelihood when the method predicts the wrong outcome with a high probability.

\begin{table*}
\centering
\caption{Comparison on expert feedback prediction task.}
\label{tab:expertfeedback}       
\begin{tabular}{lllllll}
\hline\noalign{\smallskip}
 &  \multicolumn{2}{c}{Micro-avg. AUC} &   \multicolumn{2}{c}{Accuracy} & \multicolumn{2}{c}{Average log-likelihood}  \\
 \noalign{\smallskip}\hline\noalign{\smallskip}
Method/Feedback & Length & Width & Length & Width & Length & Width   \\
\noalign{\smallskip}\hline\noalign{\smallskip}
Naive Bayes & $0.681 \pm 0.003$ & $0.716 \pm 0.006$ & $0.737 \pm 0.003$  & $\bf 0.875 \pm 0.005$ & $\bf -0.656 \pm 0.005$ & $\bf -0.395 \pm 0.004$ \\
Boosted Trees & $0.708 \pm 0.003$ &  $0.715 \pm 0.005$  & $\bf 0.748 \pm 0.009$ &  $0.872 \pm 0.003$  & $-0.746 \pm 0.028$ &  $-0.464 \pm 0.009$  \\
\textsc{SFnet} &$\bf 0.753 \pm 0.004$ & $ \bf 0.773 \pm 0.006$ &  $0.742 \pm 0.005$ & $\bf 0.876 \pm 0.003$ & $-0.698 \pm 0.011$ & $-0.409 \pm 0.007$ \\
\noalign{\smallskip}\hline
\end{tabular}
\end{table*}

\subsection{Experiments on Purchase Data}
\label{sec:2.4}
In this section, we present results on modeling customer size preferences given their purchase history. Our goal here is to predict the size of articles which customers order and keep. Note that a "customer" in this context refers to a customer account which is potentially used by multiple individuals. This is a realistic scenario for most e-commerce retail platforms and for personalized recommendation, it demonstrates the need for modeling multiple personas behind one identity. We will analyze \textsc{SFnet}'s performance on multi-user accounts in Section~\ref{sec:multiusers}.

For these experiments, we use our proprietary dataset of customer purchases spanning a period of roughly $5$ years. The purchased articles in the data belong to the sub-categories of shoes, textile and sportswear. We only consider customer accounts with at least $5$ purchases in the history. The dataset contains roughly $20$ million purchases involving around $389K$ customers and $872K$ articles. There are more than $1K$ distinct sizes in the data. Multidimensional sizes such as jeans size $30 \times 32$ and $30 \times 33$ are taken to be independent of each other. Due to overlapping size systems, a distinct size can be used in multiple clothing sub-categories.

Apart from an anonymous customer identifier, our data does not contain any other customer information. We therefore do not consider cold-start customers in this experiment\footnote{In the absence of additional features as in Table~\ref{tab:benchmarkfeats}, if (akin to Section~\ref{sec:2.2}) we learn a default customer embedding for cold-start customers, we can only expect to approximate population-level marginal distributions over kept sizes in article sub-categories, which will be rather non-informative for personalization.}.  For articles we use unique identifiers together with categorical attributes such as brand, main material, country of origin, season and $5$ taxonomical attributes which including gender (female, male or unisex), define a non-overlapping hierarchy of clothing items. 

\vspace{3mm}
\noindent{\textbf{Backtesting:}}
To simulate a realistic scenario, we perform our experiments in a backtesting setup. To that end, we split the data chronologically into train, validation and test sets. This implies that our training instances come from the past, while validation and test splits contain more recent purchases with test split containing the latest ones. In backtesting, aside from encountering cold-start customers, we may also encounter new articles in the test for which have not learned any dedicated input embeddings during training; nonetheless the default article embedding (as described in Section \ref{sec:2.2}) together with shared attributes such as brand, material, etc. allow us to evaluate new articles in the test (and validation) data split. 

With $80\%$ train, $10\%$ validation\footnote{Since the methods we compare with in this section do not require extensive hyperparameter tuning, we merge the validation split into the training data for those methods.} and $10\%$ test, we keep data split ratios the same as before. During test, we truncate and renormalize the output distributions of \textsc{SFnet} and compared methods to the available sizes of test articles. Moreover, since we allow customers to order more than one sizes, we further report top-2 and top-3 accuracies with the other performance metrics.

\vspace{3mm}
\noindent{\textbf{Bayesian Model:}}
We benchmark our approach against a recently introduced Bayesian method for size recommendation \cite{guigoures2018hierarchical}. The approach is based on a hierarchical Bayesian model exploiting the customer purchase history to learn the usual size of multiple users of a single account. Originally, the method was proposed to model both returns and keeps in a customer history, but in our setting where we are only interested in modeling size distribution of kept articles in customer accounts. In this case, the model proposed by \cite{guigoures2018hierarchical} reduces to an infinite Gaussian mixture model with an associated truncated Dirichlet process of level four (we refer to \cite{guigoures2018hierarchical} for more details). 


We train an independent instance of the Bayesian model for articles of all genders (i.e. female, male and unisex) within each of the main clothing categories in data -- including shoes and upper and lower body garments. Moreover, since the approach is meant to be for continuous size systems, we employ expert knowledge to convert alpha-numeric sizes (e.g., Small, Medium, Large) into a continuous size range. To disambiguate overlapping numerical size systems, we further use a semi-supervised Gaussian Expectation Maximization algorithm \cite{basu2002} to cluster articles based on the characteristics of their size systems (e.g., minimum, maximum and median sizes, step between sizes, etc.). Once clustered, the size that represents a cluster is defined by a domain expert.

\vspace{3mm}
\noindent{\textbf{Baseline:}}
We also estimate a population-level marginal distribution of kept sizes, which we obtain by training the Bayesian model for each clothing category and gender across all customers.



\vspace{3mm}
\noindent{\textbf{Results:}}
As shown in Table \ref{tab:all}, \textsc{SFnet} outperforms both Bayesian and baseline approaches on all the metrics. We further observe a narrowing gap between \textsc{SFnet} and the Bayesian approach w.r.t. top-$k$ accuracies. This is due to the fact that for a given article, there are usually a handful of sizes to choose from, hence increasing $k$ significantly boosts the chances of hitting the right size for both the algorithms. 

\begin{table}
\centering
\caption{Comparison on test data containing articles in various clothing categories and overlappting size systems.}
\label{tab:all}
\resizebox{\columnwidth}{!}{
\begin{tabular}{llllll}
\hline\noalign{\smallskip}
 & \multicolumn{1}{c} {Micro-avg.} &  \multicolumn{3}{c}{Accuracy}  & \multicolumn{1}{c} {Average}  \\
Method & \multicolumn{1}{c} { AUC} &  \multicolumn{1}{c}{top-1} & \multicolumn{1}{c}{top-2} & \multicolumn{1}{c}{top-3}  & \multicolumn{1}{c}{log-likelihood}  \\
\noalign{\smallskip}\hline\noalign{\smallskip}

\multicolumn{1}{l}{Baseline} &  $0.690$ & $0.24$ & $0.45$  & $0.64$ & $-1.82$\\
\multicolumn{1}{l}{Bayesian} &  $0.834$ & $0.503$ & $0.770$  & $0.886$ &  $-1.37$\\
\multicolumn{1}{l}{\textsc{SFnet}} & $\bf 0.861$  & $\bf 0.555$ & $\bf 0.795$ & $\bf 0.898$ &  $\bf -1.19$\\
\noalign{\smallskip}\hline
\end{tabular}
}
\end{table}

\subsubsection{\textbf{Dealing with Category Cold-Starts:}}
An appealing use-case for size recommendation in e-commerce fashion retail is that of category cold-start where an existing customer with purchase history in other categories orders an article from a new category. Note that for category cold-starts, the Bayesian approach defaults to the baseline approach, which is a category and gender-conditioned marginal distribution of purchased sizes. 

\vspace{3mm}
\noindent{\textbf{Results:}}
While the baseline approach recommends among available sizes, the most purchased size of a category cold-start article, we expect \textsc{SFnet} to be better than that.
Indeed tn table \ref{tab:cat-cold-start}, we find \textsc{SFnet}'s performance on cold-start recommendation in three different categories significantly better than the baseline default mode of the Bayesian approach.
\begin{table}
\centering
\caption{Category cold-start performance in three different categories.}
\label{tab:cat-cold-start}
\resizebox{\columnwidth}{!}{
\begin{tabular}{llllll}
\hline\noalign{\smallskip}
 & \multicolumn{1}{c} {Micro-avg.} &  \multicolumn{3}{c}{Accuracy}  & \multicolumn{1}{c} {Average}  \\
 & \multicolumn{1}{c} { AUC} &  \multicolumn{1}{c}{top-1} & \multicolumn{1}{c}{top-2} & \multicolumn{1}{c}{top-3}  & \multicolumn{1}{c}{log-likelihood}  \\
\noalign{\smallskip}\hline\noalign{\smallskip}
\multicolumn{6}{c} {Men's Shirts} \\
\noalign{\smallskip}\hline
\multicolumn{1}{l}{Baseline} & $0.63$ & $0.21$ & $0.44$ & $0.61$ & $-1.78$\\
\multicolumn{1}{l}{\textsc{SFnet}} & $\bf 0.723$  & $\bf 0.403$  & $\bf 0.698$  & $\bf 0.810$  & $\bf -1.63 $ \\
\noalign{\smallskip}\hline
\multicolumn{6}{c} {Jeans} \\
\noalign{\smallskip}\hline
\multicolumn{1}{l}{Baseline}  & $0.68$ & $0.20$ & $0.38$ & $0.53$ & $\bf -2.10$ \\
\multicolumn{1}{l}{\textsc{SFnet}}  &  $\bf 0.775$ & $\bf 0.295$  & $\bf 0.509$ &  $\bf 0.646$ & $ -2.22 $\\
\noalign{\smallskip}\hline
\multicolumn{6}{c} {Shoes} \\
\noalign{\smallskip}\hline
\multicolumn{1}{l}{Baseline} & $0.71$ & $0.24$ & $0.45$ & $0.62$ & $\bf -1.88$\\
\multicolumn{1}{l}{\textsc{SFnet}}  & $\bf 0.745$ & $\bf 0.293$  & $\bf 0.516$ &  $\bf 0.679$ &  $-1.99 $  \\
\noalign{\smallskip}\hline
\end{tabular}
}
\end{table}

\subsubsection{\textbf{Modeling Multiple Users Behind One Identity:}}
\label{sec:multiusers}
In our last experiment we asses \textsc{SFnet}'s capacity to deal with multiple users  behind one account. We use gender profiles (i.e. female, male or unisex) of purchased articles to assume customer accounts to be single or multi-user. Based on the gender profiles, we first filter the data to contain only those accounts with both female and male articles in the test split. We then perform ablations by partitioning the filtered accounts w.r.t. their gender distribution in the training data, yielding the three rows of Table \ref{tab:multi-users}. The first row represents user accounts that either contain female and unisex, or male and unisex articles in their training histories. During test, as indicated by male and female columns of the table, those accounts are tested on the articles of gender that was lacking in their training histories. We term such cases as `gender cold-starts'. The second row of the table represents the opposite of the first row, where accounts with female and unisex (respectively male and unisex) articles in training data are tested on female (respectively male) articles. The last row represents the accounts which contain all the three genders in their training histories and we test their performance on female vs. male articles. 
\begin{table}
    \centering
    \caption{Top-1 accuracy on multi-user accounts in test. Rows represent different types of customer histories encountered during training.}
    \label{tab:multi-users}
    \begin{tabular}{l|ll|ll|ll}
        & \multicolumn{2}{|c|}{Baseline} & \multicolumn{2}{c|}{Bayesian} & \multicolumn{2}{c}{\textsc{SFnet}}  \\
        gender & \multicolumn{1}{|c}{male} & \multicolumn{1}{c|}{female} & \multicolumn{1}{c}{male} & \multicolumn{1}{c|}{female} & \multicolumn{1}{c}{male} & \multicolumn{1}{c}{female}  \\ 
        \noalign{\smallskip}\hline
        \multicolumn{1}{l|}{cold-start} & \multicolumn{1}{c}{$0.219$} & \multicolumn{1}{c|}{$0.253$} & \multicolumn{1}{c}{$0.219$} & \multicolumn{1}{c|}{$0.253$} & \multicolumn{1}{c}{$\bf 0.325$} & \multicolumn{1}{c}{$\bf 0.300$} \\
        \multicolumn{1}{l|}{consistent} & \multicolumn{1}{c}{$0.220$} & \multicolumn{1}{c|}{$0.253$} & \multicolumn{1}{c}{$0.434$} & \multicolumn{1}{c|}{$0.496$} & \multicolumn{1}{c}{$\bf 0.496$} & \multicolumn{1}{c}{$\bf 0.559$}  \\
        \multicolumn{1}{l|}{mixed} & \multicolumn{1}{c}{$0.218$} & \multicolumn{1}{c|}{$0.253$} & \multicolumn{1}{c}{$0.396$} & \multicolumn{1}{c|}{$0.503$} & \multicolumn{1}{c}{$\bf 0.481$} & \multicolumn{1}{c}{$\bf 0.549$} \\ 
        \noalign{\smallskip}\hline
    \end{tabular}
\end{table}

\vspace{3mm}
\noindent{\textbf{Results:}}
As the Bayesian approach defaults to the baseline for the gender cold-starts, we see identical numbers for both methods in the first row of Table \ref{tab:multi-users}; to our surprise however, \textsc{SFnet}'s performance for the gender cold-starts is significantly better than the baseline marginals. We hypothesize that \textsc{SFnet} makes use of higher-order correlations discovered from multi-user accounts to achieve the results. In the second row of the table, we see \textsc{SFnet} is most accurate with user accounts that are consistently one gender (plus unisex) during training and test. For multi-user accounts in the third row, we observe a reduction in \textsc{SFnet}'s performance, yet the accuracy is significantly higher than the Bayesian (and baseline) approach. The results are indicative of \textsc{SFnet}'s capacity for modeling multiple users, although further analysis is warranted to assess~\textsc{SFnet}'s ability to disambiguate multiple intents.   

\section{Conclusion}
\label{sec:5}
In this work we proposed \textsc{SFnet}, a deep learning based methodology which combines collaborative and content-based modeling techniques to learn input and latent representations of customers and articles for size and fit prediction. The method is highly scalable and works end-to-end without requiring a priori knowledge about its prediction targets underlying ordinal structure. As demonstrated by competitive empirical performance in a variety of experiments on multiple datasets, our \textsc{SFnet} architecture offers both the flexibility and the capacity for capturing higher-order abstractions of size and fit relevant information from arbitrary customer and article features. Future extensions of this work can include multi-view objectives~\cite{Elkahky2015} (such as predicting both categorical and ordinal targets) or time-dependent modeling of customer behavior~\cite{DonkersEtAl2017} with respect to size and fit. 

\begin{acks}
We acknowledge and appreciate constructive feedback from our reviewers and area chair. We thank Alan Akbik and Calvin Seward for their valuable feedback in the preparation of this manuscript. We would also like to thank Julia Lasserre for helpful discussions on the design of experiments on customer purchase data.
\end{acks}

\bibliographystyle{ACM-Reference-Format}
\bibliography{biblio.bib}   


\begin{thebibliography}{25}


\ifx \showCODEN    \undefined \def \showCODEN     #1{\unskip}     \fi
\ifx \showDOI      \undefined \def \showDOI       #1{#1}\fi
\ifx \showISBNx    \undefined \def \showISBNx     #1{\unskip}     \fi
\ifx \showISBNxiii \undefined \def \showISBNxiii  #1{\unskip}     \fi
\ifx \showISSN     \undefined \def \showISSN      #1{\unskip}     \fi
\ifx \showLCCN     \undefined \def \showLCCN      #1{\unskip}     \fi
\ifx \shownote     \undefined \def \shownote      #1{#1}          \fi
\ifx \showarticletitle \undefined \def \showarticletitle #1{#1}   \fi
\ifx \showURL      \undefined \def \showURL       {\relax}        \fi
\providecommand\bibfield[2]{#2}
\providecommand\bibinfo[2]{#2}
\providecommand\natexlab[1]{#1}
\providecommand\showeprint[2][]{arXiv:#2}

\bibitem[\protect\citeauthoryear{Abdulla and Borar}{Abdulla and Borar}{2017}]%
        {abdulla2017}
\bibfield{author}{\bibinfo{person}{G~Mohammed Abdulla} {and}
  \bibinfo{person}{Sumit Borar}.} \bibinfo{year}{2017}\natexlab{}.
\newblock \showarticletitle{Size recommendation system for fashion e-commerce}.
  In \bibinfo{booktitle}{\emph{Workshop on Machine Learning Meets Fashion,
  KDD}}.
\newblock


\bibitem[\protect\citeauthoryear{Basu, Banerjee, and Mooney}{Basu
  et~al\mbox{.}}{2002}]%
        {basu2002}
\bibfield{author}{\bibinfo{person}{Sugato Basu}, \bibinfo{person}{Arindam
  Banerjee}, {and} \bibinfo{person}{Raymond Mooney}.}
  \bibinfo{year}{2002}\natexlab{}.
\newblock \showarticletitle{Semi-supervised clustering by seeding}. In
  \bibinfo{booktitle}{\emph{In Proceedings of 19th International Conference on
  Machine Learning}}.
\newblock


\bibitem[\protect\citeauthoryear{Bromley, Guyon, LeCun, S\"{a}ckinger, and
  Shah}{Bromley et~al\mbox{.}}{1994}]%
        {BromleyEtAl1993}
\bibfield{author}{\bibinfo{person}{Jane Bromley}, \bibinfo{person}{Isabelle
  Guyon}, \bibinfo{person}{Yann LeCun}, \bibinfo{person}{Eduard S\"{a}ckinger},
  {and} \bibinfo{person}{Roopak Shah}.} \bibinfo{year}{1994}\natexlab{}.
\newblock \showarticletitle{Signature Verification using a "Siamese" time delay
  neural network}.
\newblock In \bibinfo{booktitle}{\emph{Advances in Neural Information
  Processing Systems 6}}, \bibfield{editor}{\bibinfo{person}{J.~D. Cowan},
  \bibinfo{person}{G.~Tesauro}, {and} \bibinfo{person}{J.~Alspector}} (Eds.).
  \bibinfo{pages}{737--744}.
\newblock


\bibitem[\protect\citeauthoryear{Donkers, Loepp, and Ziegler}{Donkers
  et~al\mbox{.}}{2017}]%
        {DonkersEtAl2017}
\bibfield{author}{\bibinfo{person}{Tim Donkers}, \bibinfo{person}{Benedikt
  Loepp}, {and} \bibinfo{person}{J\"{u}rgen Ziegler}.}
  \bibinfo{year}{2017}\natexlab{}.
\newblock \showarticletitle{Sequential user-based recurrent neural network
  recommendations}. In \bibinfo{booktitle}{\emph{Proceedings of the 11th ACM
  Conference on Recommender Systems}} \emph{(\bibinfo{series}{RecSys '17})}.
\newblock


\bibitem[\protect\citeauthoryear{Elkahky, Song, and He}{Elkahky
  et~al\mbox{.}}{2015}]%
        {Elkahky2015}
\bibfield{author}{\bibinfo{person}{Ali~Mamdouh Elkahky}, \bibinfo{person}{Yang
  Song}, {and} \bibinfo{person}{Xiaodong He}.} \bibinfo{year}{2015}\natexlab{}.
\newblock \showarticletitle{A multi-view deep learning approach for cross
  domain user modeling in recommendation systems}. In
  \bibinfo{booktitle}{\emph{Proceedings of the 24th International Conference on
  World Wide Web}}. International World Wide Web Conferences Steering
  Committee, \bibinfo{pages}{278--288}.
\newblock


\bibitem[\protect\citeauthoryear{Graham, Knuth, Patashnik, and Liu}{Graham
  et~al\mbox{.}}{1989}]%
        {graham1989concrete}
\bibfield{author}{\bibinfo{person}{Ronald~L Graham}, \bibinfo{person}{Donald~E
  Knuth}, \bibinfo{person}{Oren Patashnik}, {and} \bibinfo{person}{Stanley
  Liu}.} \bibinfo{year}{1989}\natexlab{}.
\newblock \showarticletitle{Concrete mathematics: a foundation for computer
  science}.
\newblock \bibinfo{journal}{\emph{Computers in Physics}} \bibinfo{volume}{3},
  \bibinfo{number}{5} (\bibinfo{year}{1989}), \bibinfo{pages}{106--107}.
\newblock


\bibitem[\protect\citeauthoryear{Guigour{\`e}s, Ho, Koriagin, Sheikh, Bergmann,
  and Shirvany}{Guigour{\`e}s et~al\mbox{.}}{2018}]%
        {guigoures2018hierarchical}
\bibfield{author}{\bibinfo{person}{Romain Guigour{\`e}s},
  \bibinfo{person}{Yuen~King Ho}, \bibinfo{person}{Evgenii Koriagin},
  \bibinfo{person}{Abdul-Saboor Sheikh}, \bibinfo{person}{Urs Bergmann}, {and}
  \bibinfo{person}{Reza Shirvany}.} \bibinfo{year}{2018}\natexlab{}.
\newblock \showarticletitle{A hierarchical bayesian model for size
  recommendation in fashion}. In \bibinfo{booktitle}{\emph{Proceedings of the
  12th ACM Conference on Recommender Systems}}. ACM, \bibinfo{pages}{392--396}.
\newblock


\bibitem[\protect\citeauthoryear{He, Zhang, Ren, and Sun}{He
  et~al\mbox{.}}{2016}]%
        {resnet2016}
\bibfield{author}{\bibinfo{person}{Kaiming He}, \bibinfo{person}{Xiangyu
  Zhang}, \bibinfo{person}{Shaoqing Ren}, {and} \bibinfo{person}{Jian Sun}.}
  \bibinfo{year}{2016}\natexlab{}.
\newblock \showarticletitle{Deep residual learning for image recognition}. In
  \bibinfo{booktitle}{\emph{Proceedings of the IEEE conference on computer
  vision and pattern recognition}}. \bibinfo{pages}{770--778}.
\newblock


\bibitem[\protect\citeauthoryear{He, Liao, Zhang, Nie, Hu, and Chua}{He
  et~al\mbox{.}}{2017}]%
        {ncf2017}
\bibfield{author}{\bibinfo{person}{Xiangnan He}, \bibinfo{person}{Lizi Liao},
  \bibinfo{person}{Hanwang Zhang}, \bibinfo{person}{Liqiang Nie},
  \bibinfo{person}{Xia Hu}, {and} \bibinfo{person}{Tat-Seng Chua}.}
  \bibinfo{year}{2017}\natexlab{}.
\newblock \showarticletitle{Neural collaborative filtering}. In
  \bibinfo{booktitle}{\emph{Proceedings of the 26th International Conference on
  World Wide Web}}. International World Wide Web Conferences Steering
  Committee, \bibinfo{pages}{173--182}.
\newblock


\bibitem[\protect\citeauthoryear{Huang, He, Gao, Deng, Acero, and Heck}{Huang
  et~al\mbox{.}}{2013}]%
        {DSSM2013}
\bibfield{author}{\bibinfo{person}{Po-Sen Huang}, \bibinfo{person}{Xiaodong
  He}, \bibinfo{person}{Jianfeng Gao}, \bibinfo{person}{Li Deng},
  \bibinfo{person}{Alex Acero}, {and} \bibinfo{person}{Larry Heck}.}
  \bibinfo{year}{2013}\natexlab{}.
\newblock \showarticletitle{Learning deep structured semantic models for web
  search using clickthrough Data}.
\newblock \bibinfo{journal}{\emph{ACM SIGKDD Explorations Newsletter}}
  \bibinfo{volume}{3}, \bibinfo{number}{1}, \bibinfo{pages}{27--32}.
\newblock


\bibitem[\protect\citeauthoryear{Ioffe and Szegedy}{Ioffe and Szegedy}{2015}]%
        {ioffe2015}
\bibfield{author}{\bibinfo{person}{Sergey Ioffe} {and}
  \bibinfo{person}{Christian Szegedy}.} \bibinfo{year}{2015}\natexlab{}.
\newblock \showarticletitle{Batch normalization: Accelerating deep network
  training by reducing internal covariate shift}.
\newblock \bibinfo{journal}{\emph{arXiv preprint arXiv:1502.03167}}
  (\bibinfo{year}{2015}).
\newblock


\bibitem[\protect\citeauthoryear{Johnson}{Johnson}{2014}]%
        {johnson2014logistic}
\bibfield{author}{\bibinfo{person}{Christopher~C Johnson}.}
  \bibinfo{year}{2014}\natexlab{}.
\newblock \showarticletitle{Logistic matrix factorization for implicit feedback
  data}.
\newblock \bibinfo{journal}{\emph{Advances in Neural Information Processing
  Systems}}  \bibinfo{volume}{27} (\bibinfo{year}{2014}).
\newblock


\bibitem[\protect\citeauthoryear{Kingma and Ba}{Kingma and Ba}{2014}]%
        {kingma2014}
\bibfield{author}{\bibinfo{person}{Diederik~P Kingma} {and}
  \bibinfo{person}{Jimmy Ba}.} \bibinfo{year}{2014}\natexlab{}.
\newblock \showarticletitle{Adam: A method for stochastic optimization}.
\newblock \bibinfo{journal}{\emph{arXiv preprint arXiv:1412.6980}}
  (\bibinfo{year}{2014}).
\newblock


\bibitem[\protect\citeauthoryear{Koren and Bell}{Koren and Bell}{2015}]%
        {KorenBell2015}
\bibfield{author}{\bibinfo{person}{Yehuda Koren} {and}
  \bibinfo{person}{Robert~M. Bell}.} \bibinfo{year}{2015}\natexlab{}.
\newblock \showarticletitle{Advances in Collaborative Filtering}.
\newblock In \bibinfo{booktitle}{\emph{Recommender Systems Handbook}}.
  \bibinfo{pages}{77--118}.
\newblock


\bibitem[\protect\citeauthoryear{Manning, Raghavan, and Sch{\"{u}}tze}{Manning
  et~al\mbox{.}}{2008}]%
        {manning2013introduction}
\bibfield{author}{\bibinfo{person}{Christopher~D. Manning},
  \bibinfo{person}{Prabhakar Raghavan}, {and} \bibinfo{person}{Hinrich
  Sch{\"{u}}tze}.} \bibinfo{year}{2008}\natexlab{}.
\newblock \bibinfo{booktitle}{\emph{Introduction to information retrieval}}.
\newblock \bibinfo{publisher}{Cambridge University Press}.
\newblock
\showISBNx{978-0-521-86571-5}


\bibitem[\protect\citeauthoryear{Micci-Barreca}{Micci-Barreca}{2001}]%
        {micci2001preprocessing}
\bibfield{author}{\bibinfo{person}{Daniele Micci-Barreca}.}
  \bibinfo{year}{2001}\natexlab{}.
\newblock \showarticletitle{A preprocessing scheme for high-cardinality
  categorical attributes in classification and prediction problems}.
\newblock \bibinfo{journal}{\emph{ACM SIGKDD Explorations Newsletter}}
  \bibinfo{volume}{3}, \bibinfo{number}{1} (\bibinfo{year}{2001}),
  \bibinfo{pages}{27--32}.
\newblock


\bibitem[\protect\citeauthoryear{Mikolov, Chen, Corrado, and Dean}{Mikolov
  et~al\mbox{.}}{2018}]%
        {word2vec}
\bibfield{author}{\bibinfo{person}{Tomas Mikolov}, \bibinfo{person}{Kai Chen},
  \bibinfo{person}{Greg Corrado}, {and} \bibinfo{person}{Jeffrey Dean}.}
  \bibinfo{year}{2018}\natexlab{}.
\newblock \showarticletitle{Efficient estimation of word representations in
  vector space}.
\newblock \bibinfo{journal}{\emph{Workshop in International Conference on
  Learning Representations (ICLR)}} (\bibinfo{year}{2018}).
\newblock


\bibitem[\protect\citeauthoryear{Misra, Wan, and McAuley}{Misra
  et~al\mbox{.}}{2018}]%
        {misra2018decomposing}
\bibfield{author}{\bibinfo{person}{Rishabh Misra}, \bibinfo{person}{Mengting
  Wan}, {and} \bibinfo{person}{Julian McAuley}.}
  \bibinfo{year}{2018}\natexlab{}.
\newblock \showarticletitle{Decomposing fit semantics for product size
  recommendation in metric spaces}. In \bibinfo{booktitle}{\emph{Proceedings of
  the 12th ACM Conference on Recommender Systems}}. ACM,
  \bibinfo{pages}{422--426}.
\newblock


\bibitem[\protect\citeauthoryear{Peng and Mouhannad}{Peng and
  Mouhannad}{2014}]%
        {PengSayegh2014}
\bibfield{author}{\bibinfo{person}{Fanke Peng} {and}
  \bibinfo{person}{Al-Sayegh. Mouhannad}.} \bibinfo{year}{2014}\natexlab{}.
\newblock \showarticletitle{Personalised Size Recommendation for Online
  Fashion}. In \bibinfo{booktitle}{\emph{6th International conference on mass
  customization and personalization in Central Europe}}. \bibinfo{pages}{1--6}.
\newblock


\bibitem[\protect\citeauthoryear{Pisut and Connell}{Pisut and Connell}{2007}]%
        {Pisut2017}
\bibfield{author}{\bibinfo{person}{Gina Pisut} {and} \bibinfo{person}{Lenda~Jo
  Connell}.} \bibinfo{year}{2007}\natexlab{}.
\newblock \showarticletitle{Fit preferences of female consumers in the USA}.
\newblock \bibinfo{journal}{\emph{Journal of Fashion Marketing and Management:
  An International Journal}} \bibinfo{volume}{11}, \bibinfo{number}{3}
  (\bibinfo{year}{2007}), \bibinfo{pages}{366--379}.
\newblock


\bibitem[\protect\citeauthoryear{Sembium, Rastogi, Saroop, and Merugu}{Sembium
  et~al\mbox{.}}{2017}]%
        {sembium2017}
\bibfield{author}{\bibinfo{person}{Vivek Sembium}, \bibinfo{person}{Rajeev
  Rastogi}, \bibinfo{person}{Atul Saroop}, {and} \bibinfo{person}{Srujana
  Merugu}.} \bibinfo{year}{2017}\natexlab{}.
\newblock \showarticletitle{Recommending product sizes to customers}. In
  \bibinfo{booktitle}{\emph{Proceedings of the 11th ACM Conference on
  Recommender Systems}}. ACM, \bibinfo{pages}{243--250}.
\newblock


\bibitem[\protect\citeauthoryear{Sembium, Rastogi, Tekumalla, and
  Saroop}{Sembium et~al\mbox{.}}{2018}]%
        {Sembium2018}
\bibfield{author}{\bibinfo{person}{Vivek Sembium}, \bibinfo{person}{Rajeev
  Rastogi}, \bibinfo{person}{Lavanya Tekumalla}, {and} \bibinfo{person}{Atul
  Saroop}.} \bibinfo{year}{2018}\natexlab{}.
\newblock \showarticletitle{Bayesian models for product size recommendations}.
  In \bibinfo{booktitle}{\emph{Proceedings of the 25th International Conference
  on World Wide Web}}. \bibinfo{pages}{679--687}.
\newblock


\bibitem[\protect\citeauthoryear{Shi, Larson, and Hanjalic}{Shi
  et~al\mbox{.}}{2014}]%
        {ShiEtAl2014}
\bibfield{author}{\bibinfo{person}{Yue Shi}, \bibinfo{person}{Martha Larson},
  {and} \bibinfo{person}{Alan Hanjalic}.} \bibinfo{year}{2014}\natexlab{}.
\newblock \showarticletitle{Collaborative filtering beyond the user-item
  matrix: A survey of the state-of-the-art and future challenges}.
\newblock \bibinfo{journal}{\emph{ACM Comput. Surv.}} (\bibinfo{year}{2014}).
\newblock


\bibitem[\protect\citeauthoryear{Socher, Chen, Manning, and Ng}{Socher
  et~al\mbox{.}}{2013}]%
        {socher2013}
\bibfield{author}{\bibinfo{person}{Richard Socher}, \bibinfo{person}{Danqi
  Chen}, \bibinfo{person}{Christopher~D Manning}, {and} \bibinfo{person}{Andrew
  Ng}.} \bibinfo{year}{2013}\natexlab{}.
\newblock \showarticletitle{Reasoning with neural tensor networks for knowledge
  base completion}. In \bibinfo{booktitle}{\emph{Advances in neural information
  processing systems}}. \bibinfo{pages}{926--934}.
\newblock


\bibitem[\protect\citeauthoryear{Srivastava, Hinton, Krizhevsky, Sutskever, and
  Salakhutdinov}{Srivastava et~al\mbox{.}}{2014}]%
        {srivastava2014}
\bibfield{author}{\bibinfo{person}{Nitish Srivastava},
  \bibinfo{person}{Geoffrey Hinton}, \bibinfo{person}{Alex Krizhevsky},
  \bibinfo{person}{Ilya Sutskever}, {and} \bibinfo{person}{Ruslan
  Salakhutdinov}.} \bibinfo{year}{2014}\natexlab{}.
\newblock \showarticletitle{Dropout: a simple way to prevent neural networks
  from overfitting}.
\newblock \bibinfo{journal}{\emph{The journal of machine learning research}}
  \bibinfo{volume}{15}, \bibinfo{number}{1} (\bibinfo{year}{2014}),
  \bibinfo{pages}{1929--1958}.
\newblock


\end{thebibliography}

\end{document}